\documentclass[a4paper,fleqn]{cas-dc}

\usepackage[authoryear,longnamesfirst]{natbib}

\def\tsc#1{\csdef{#1}{\textsc{\lowercase{#1}}\xspace}}
\tsc{WGM}
\tsc{QE}

\usepackage{algorithm}      
\usepackage{algorithmic}

\begin{document}
\let\WriteBookmarks\relax
\def\floatpagepagefraction{1}
\def\textpagefraction{.001}

\shorttitle{}    

\shortauthors{}  

\title [mode = title]{Width Expansion as a Method for Class Incremental Learning} 



%

\author[1]{André L. S. Conde}[orcid=0000-0002-5227-5271]

\cormark[1]


\ead{andre.conde@unesp.br}


\credit{Writing – original draft, Visualization, Methodology, Investigation, Software, Formal analysis, Conceptualization}

\affiliation[1]{organization={Institute of Geosciences and Exact Sciences, São Paulo State University (UNESP)},
            addressline={Av. 24-A, 1515}, 
            city={Rio Claro},
            postcode={13506-692}, 
            state={SP},
            country={Brazil}}

\author[2]{Yehia Elkhatib}[orcid=0000-0003-4639-436X]
\credit{review \& editing, Investigation, Validation}

\ead{yehia.elkhatib@glasgow.ac.uk}

\author[1]{Cateano M. Ranieri}[orcid=0000-0001-5680-9085]


\ead{cm.ranieri@unesp.br}


\credit{review \& editing, Investigation, Project administration, Supervision, Funding acquisition}

\affiliation[2]{organization={School of Computing Science, University of Glasgow},
            city={Glasgow},
            postcode={G12 8QQ}, 
            state={Scotland},
            country={United Kingdom}}

\cortext[1]{Corresponding author}



\begin{abstract}
Class Incremental Learning (Class-IL) is a critical frontier in Continual Learning, which requires that models learn new classes over time while preserving previously acquired knowledge without access to past data or task identity. This setting acutely intensifies the stability-plasticity dilemma, making catastrophic forgetting a central challenge.
Existing approaches to mitigate forgetting can be broadly categorized into (i) regularization-based methods, which constrain parameter updates; (ii) functional methods, which rely on knowledge distillation; (iii) replay-based strategies, which revisit stored samples; and (iv) architectural approaches, which expand model capacity over time.
Regarding architectural expansion, most methods rely on an explicit task identifier or predefined growth strategies, which limit their applicability in Class-IL settings where task boundaries are not available during inference.
To address this challenge, this work proposes a dynamic width expansion method that increases the number of neurons within existing layers based on a normalized loss criterion, eliminating the need for task-specific information. Additionally, an attention mechanism with persistent key-value memory is incorporated to stabilize feature representations and mitigate interference between previously learned and newly introduced classes.
The proposed approach is evaluated on Split MNIST and Split CIFAR-100 benchmarks under the standard Class-IL protocol. Experiments compare fixed-capacity models with dynamically expanding architectures, both in isolation and in combination with the attention mechanism, using established continual learning strategies such as EWC, LwF, and A-GEM.
The results suggest that progressively increasing the model capacity leads to consistent improvements over fixed architectures, particularly when combined with functional methods and A-GEM. The combination of width expansion and attention mechanisms yields the most consistent gains.
In conclusion, a dynamically expanding network width based on representational demand provides an effective and flexible strategy for Class-IL. However, caution is required, as uncontrolled growth may lead to overfitting and increased computational cost. 
\end{abstract}





\maketitle

\section{Introduction} \label{Introduction}
Deep neural networks have achieved remarkable performance across a wide range of tasks when trained under the traditional assumption of independent and identically distributed (i.i.d) data \citep{cao2022beyond}. However, in many real-world applications, data are not available all at once but arrive sequentially over time, a problem with non-stationary data that remains unsolved \citep{CLDNNRao:2020}. In such scenarios, models are required to continuously incorporate new information while retaining previously acquired knowledge. This setting is commonly referred to as continual or incremental learning. A central challenge in this paradigm is \textit{catastrophic forgetting}, in which learning new information leads to significant performance degradation on previously learned tasks or classes \citep{liu_incremental_2023}. 

Among the different continual learning scenarios, class-incremental learning (Class-IL) stands out as one of the most challenging. In this setting, the model must learn to discriminate among an ever-growing set of classes using a single shared classifier, without access to task identity during inference. As new classes are introduced incrementally, the model must balance preserving prior knowledge with acquiring new information, thereby intensifying the well-known stability-plasticity dilemma \citep{kim2023stability}.

A broad range of methods has been proposed to address catastrophic forgetting in Class-IL \citep{van2022three}. Regularization-based approaches constrain parameter updates to protect knowledge deemed important for previously learned classes. Functional methods, by contrast, aim to preserve the model's input-output mapping via knowledge distillation. Replay-based strategies mitigate forgetting by revisiting past data, either through stored samples or generative models. More recently, architectural approaches based on network expansion have been explored, increasing model capacity as new tasks or classes are introduced. Template-Based methods, in turn, retain compact representations of past classes in the form of prototypes or exemplars to guide future predictions. Finally, model rectification approaches aim to correct the bias introduced during incremental updates, typically by re-balancing decision boundaries or calibrating classifier outputs \citep{CILSurvey:2024}.

While these approaches have shown promising results, they also present limitations. Regularization and functional methods may excessively restrict model plasticity, replay-based methods introduce additional memory or computational overhead, and expansion-based approaches often rely on task-specific components or predefined growth strategies. Template-based methods depend on the quality and representativeness of stored prototypes, and model rectification techniques may require additional calibration steps or assumptions about data distribution shifts. \citep{CILSurvey:2024}

Recent work highlights the growing momentum of expansion-based methods in the CIL literature. Approaches such as RNE \citep{jiang_recurrent_2026} and Orth-DER \citep{dong_dynamic_2026} extend this paradigm through recurrent inter-expert connections and orthogonality, respectively, while applications to open-world streaming \citep{li_insertion_2026} and real-time surveillance \citep{hussain_class-incremental_2026} demonstrate its reach across diverse settings. Nevertheless, most of these methods increase model capacity by adding new task-specific modules, which introduces dependence on task identifiers at inference or leads to accelerated parameter growth. In this work, we propose a complementary strategy that expands representational capacity within existing layers, preserving the single shared classifier required by the standard Class-IL protocol.

To further reduce interference between previously learned and newly introduced classes, we incorporate a linear attention mechanism augmented with a persistent key-value memory. This mechanism provides a stable reference across incremental steps, helping to mitigate catastrophic forgetting and representational drift, while enabling the model to adaptively maintain its representations to new data.

The main contributions of this work are:
\begin{itemize}
    \item A dynamic width expansion mechanism for Class-IL (Section \ref{Expansion_Mechanism}): a task-agnostic method that increases the number of neurons within existing layers based on a normalized loss criterion, eliminating the need for explicit task identifiers during inference.
    \item A linear attention mechanism with persistent key-value memory (Section \ref{Attention_Mechanism}): a complementary module that stabilizes feature representations across incremental steps and mitigates interference between previously learned and newly introduced classes.
    \item A systematic empirical evaluation (Sections \ref{Experimental_Setup} and \ref{Results}): experiments on Split MNIST and Split CIFAR-100 benchmarks compare fixed-capacity and dynamically expanding architectures, in isolation and in combination with the attention mechanism, across established continual learning strategies including EWC, LwF, and A-GEM.
\end{itemize}


\section{Related Work}\label{Related_Works}
Catastrophic forgetting in Class-IL has motivated a broad range of approaches, which can be grouped into six paradigms \citep{CILSurvey:2024}: regularization-based, functional, replay-based, architectural, template-based, and model rectification.

Regularization-based methods such as EWC \citep{EWCKirkpatrick:2017} and SI \citep{SIZenke:2017} constrain parameter updates by penalizing deviations from weights deemed important for prior tasks. While effective at reducing interference, these methods tend to over-restrict plasticity in Class-IL, where all classes share a common output space.

Functional methods such as LwF \citep{LwFLi:2017} and LwM \citep{LwMDhar:2019} preserve the model's input-output behavior via knowledge distillation, rather than constraining individual parameters. LwF encourages consistency with previous model outputs through a temperature-scaled KL divergence, while LwM extends this by additionally penalizing divergence in intermediate attention maps. Both approaches offer more flexibility than parameter constraints, but may become insufficient when substantial representational adaptation is required.

Replay-based methods such as ER \citep{ER:2019} and A-GEM \citep{AGEMChaudhry:2019} mitigate forgetting by revisiting stored samples from prior classes. A-GEM constrains gradient updates so that new learning does not increase loss on buffered samples. These methods often achieve strong performance, but introduce memory overhead and raise scalability concerns as the number of observed classes grows.

Architectural expansion methods increase model capacity as new classes are introduced. DER \citep{DERYan:2021} appends a new extractor while freezing the previous extractor at each step; DNE \citep{DNEHu:2023} extends this with dense cross-step connections; EASE \citep{zhou_expandable_2024} employs lightweight adapters on a frozen pre-trained backbone; and KANets \citep{fu_knowledge_2023} separates old and new knowledge into parallel branches before compressing them. Recent approaches such as RNE \citep{jiang_recurrent_2026} and Orth-DER \citep{dong_dynamic_2026} further extend this paradigm via recurrent inter-expert connections and orthogonal constraints, respectively. A common limitation across these methods is reliance on task-specific modules or explicit task identifiers, which are unavailable under the standard Class-IL protocol.

Template-based methods such as iCaRL \citep{iCaRLRebuffi:2017}, CoPE \citep{de_lange_continual_2021}, and SDC \citep{yu_semantic_2020} retain compact class prototypes to guide classification and representation learning. Their effectiveness is bounded by exemplar quality and the per-class memory budget, both of which degrade as the class space grows. Model rectification methods such as WA \citep{zhao_maintaining_2020} and FACT \citep{zhou_forward_2022} address classifier bias introduced during incremental updates, but operate primarily at the output layer and may be insufficient when feature-level drift is substantial.

The method proposed in this work addresses a gap in the architectural expansion paradigm: rather than introducing per-task modules, it expands representational capacity within existing layers, guided by a normalized loss criterion, making it  directly applicable to the standard Class-IL setting without access to task identity at inference.

\section{Proposed Approach}\label{Proposed_Approach}
The proposed method operates under the standard Class-IL setting, where a model is trained over a sequence of incremental steps, each introducing a disjoint set of new classes. At each step, the model must learn to recognize newly introduced classes while preserving knowledge of previously learned ones, using a single shared classifier and without access to explicit task identifiers during inference.

The central motivation for the proposed approach stems from the stability-plasticity dilemma inherent in this setting \cite{van2022three}. As new classes are introduced, a fixed-capacity model must reallocate resources, often at the expense of previously acquired knowledge. To address this, we propose incrementally expanding the number of neurons within existing layers, allowing the model to accommodate new classes with reduced interference to older ones. However, increasing capacity alone is insufficient to prevent representational drift, as newly added parameters may disrupt the feature space learned for prior classes. To mitigate this, we further incorporate an attention mechanism with persistent key-value memory, which stabilizes feature representations across incremental steps and reduces interference between previously learned and newly introduced classes.

In contrast to network expansion approaches such as DER and DNE, the proposed method increases representational capacity within existing layers rather than introducing new task-specific modules. While DER expands the feature space by concatenating independent extractors and DNE promotes feature reuse via cross-task attention, both approaches rely on the progressive addition of new structures tied to specific learning steps. The proposed approach, called Width Expansion (WE) and its Width Dynamic layers (WD), departs from this paradigm by enabling continuous adaptation of learned representations, thereby allowing a more flexible balance between stability and plasticity: the model can both retain prior knowledge and refine its feature space as capacity expands.

\subsection{Preliminaries and Loss Function}

The proposed method builds upon a multiclass classification objective based on the cross-entropy loss, defined as:
\begin{equation}
    \label{crossentropyeq}
    \mathcal{L}_{CE} = -\sum^{K}_{k=1}y_klog(p_k)
\end{equation}
where \(y_k\) denotes the ground-truth label and \(p_k\) is the predicted probability for class \(k\), obtained through the softmax function:
\begin{equation}
\label{softmaxeq}
    p_k = softmax(z_k) = \frac{e^{z_k}}{\sum^{C}_{j=1}e^{zj}} 
\end{equation}
with \(z_k\) representing the logit associated with class \(k\).

An important reference point in the Class-IL setting is the expected loss of a model that has not yet adapted to the classes introduced at the current incremental step. Under the assumption that such a model produces approximately uniform predictions over all observed classes, the expected cross-entropy loss is given by:
\begin{equation}
    \label{approxCE}
    \mathcal{L}_{max} = ln(|seen\_classes|)
\end{equation}
where \(|C_{seen}|\) denotes the total number of classes observed up to and including the current incremental step. This quantity serves as a scale-invariant upper bound on the loss, and is subsequently used to normalize the expansion criterion described in the following section.

\subsection{Expansion Mechanism}
\label{Expansion_Mechanism}
At the beginning of each incremental step, the model's capacity to represent newly introduced classes is assessed before any adaptation takes place. Specifically, the average cross-entropy loss is computed over the training samples of the current step using the model trained at the previous step, as described in Algorithm \ref{alg_loss_mean}. This value reflects the degree to which the current architecture can accommodate the new classes without modification.


\begin{algorithm}[!ht]
\caption{Mean Loss Computation}
\label{alg_loss_mean}
\begin{algorithmic}[1]

\STATE $L \gets 0$; \quad $N \gets 0$

\FOR{mini-batch $(x, y)$ in the dataset}
    \STATE $o \gets model(x)$
    \STATE $\ell \gets criterion(o, y)$
    \STATE $L \gets L + \ell \cdot |x|$; \quad $N \gets N + |x|$
\ENDFOR

\STATE $avg\_loss \gets L / N$
\RETURN $avg\_loss$
\end{algorithmic}
\end{algorithm}
To obtain a scale-invariant measure of representational saturation, this average loss is normalized by \(\mathcal{L}_{max}\) as defined in Equation \ref{approxCE}, yielding a normalized loss \(g\) bounded between 0 and 1, as described in Algorithm \ref{alg_global_loss}. When the average loss remains below a predefined threshold, the model is assumed to have sufficient capacity to adapt to the new classes without architectural modification, and no expansion is performed.
\begin{algorithm}[!ht]
\caption{Normalized Global Loss}
\label{alg_global_loss}
\begin{algorithmic}[1]

\REQUIRE $avg\_loss$, $C_{seen}$, $loss\_threshold$

\STATE $L_{\max} \gets \ln(|C_{seen}|)$
\STATE $g \gets avg\_loss / L_{\max}$

\STATE $g \gets \min(g, 1)$

\IF{$avg\_loss \le loss\_threshold$}
    \RETURN
\ENDIF

\RETURN $g$

\end{algorithmic}
\end{algorithm}
When expansion is required, it is carried out on selected layers using a combination of the global difficulty signal \(g\) and local neuron utilization statistics. For each width-dynamic layer (WD) \(L\), a local usage ratio \(u\) is computed, reflecting how actively the layer's neurons have been engaged during previous training steps. The expansion coefficient is then defined as:
\begin{equation}
    c = gw_{loss} + u*w_{local}
\end{equation}
where \(w_{loss}\) and \(w_{local}\) are weighting factors that control the relative influence of the global loss signal and local utilization estimate, respectively. The number of neurons to be added is proportional to the layer size and is constrained within a bounded range to prevent uncontrolled growth, as detailed in Algorithm \ref{alg_layer_expansion}.

\begin{algorithm}[!h]
\caption{Expansion per Width Dynamic Layer}
\label{alg_layer_expansion}
\begin{algorithmic}[1]

\REQUIRE $g$, $model$, $w_{local}$, $w_{loss}$, $growth\_factor$

\FOR{Dynamic Width Layer $L$ in $model$}

    \STATE $u \gets L.stats.usage\_ratio()$
    \STATE $c \gets w_{\text{loss}} \cdot g + w_{\text{local}} \cdot u$

    \STATE $b \gets L.out\_features \cdot growth\_factor$
    \STATE $n \gets round(b \cdot c)$

    \STATE $m_{\min} \gets \text{(}0.05 \cdot L.out\_features\text{ if } g>0.5\text{, else } 0\text{)}$
    \STATE $m_{\max} \gets 0.75 \cdot L.out\_features$

    \STATE $n \gets \max(m_{\min}, \min(n, m_{\max}))$

    \IF{$n > 0$}
        \STATE expand $L$ with $n$ neurons
        \STATE update connections with other layers
        \STATE update optimizer
    \ENDIF

\ENDFOR

\end{algorithmic}
\end{algorithm}

This procedure ensures that the expansion process is task-agnostic, bounded, and driven by representational demand rather than any task identifier, making it directly applicable to the Class-IL setting.

\subsection{Attention Mechanism}
\label{Attention_Mechanism}
While width expansion increases the model's representational capacity, it does not explicitly address interference that can arise between previously learned and newly introduced classes. To further stabilize representations during incremental updates, we incorporate a modified linear attention mechanism inspired by \cite{LinearAttetionKatharopoulos:2020}, augmented with a persistent key-value memory.

\begin{figure}
    \centering
    \includegraphics[width=\columnwidth]{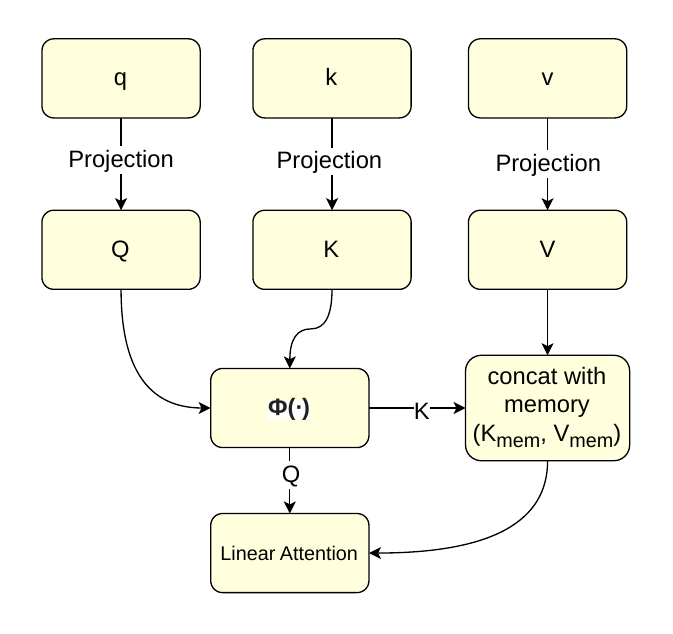}
    \caption{Attention Mechanism}
    \label{fig:Attention_Mechanism}
\end{figure}

Let \(q, \, k\) and \(v\) denote the input query, key, and value sequences. These inputs are first projected into a shared attention space through learnable linear transformations:
\begin{equation}
  Q=\phi(W_Qq),\,K=\phi(W_Kk),\,V=W_Vv  
\end{equation}
where \(W_Q, \, W_K\), and \(W_V\) are learnable projection matrices and \(\phi(\cdot)\) is a positive feature map defined as:
\begin{equation}
  \phi(x)=ELU(x)+1  
\end{equation}
where \(ELU(\cdot)\) is an exponential linear unit function \citep{clevert_fast_2016}, which enables the efficient linear attention formulation by ensuring non-negative kernel values.

To provide a stable reference across incremental learning steps, a persistent memory module is introduced, composed of learnable key-value pairs:\[k_{mem} \in \mathbb{R}^{M\times d}, \, v_{mem} \in \mathbb{R}^{M\times d}\] where \(M\) denotes the memory size and \(d\) the attention dimensionality. These parameters are initialized using orthogonal initialization to encourage diversity among stored representations.
During the forward pass, the memory vectors are concatenated with the projected keys and values: \[K'=[K;k_{mem}], \; V'=[V;v_{mem}]\]
Given the augmented key-value set, the linear attention output is computed as:
\begin{equation}
    KV=K'^TV'
\end{equation}
\begin{equation}
    Z=\frac{1}{Q(K'^T)}
\end{equation}
\begin{equation}
    Att(Q,K',V')=(Q\cdot KV)\odot Z
\end{equation}
This formulation reduces the quadratic complexity of standard self-attention to linear complexity with respect to sequence length, making it more suitable for the incremental learning setting. Notably, the persistent memory allows the model to attend to stable key-value representations consolidated during previous training stages, thereby mitigating representational drift and helping to preserve previously acquired knowledge as new classes are introduced.

\section{Experimental Setup}\label{Experimental_Setup}
The experiments in this section are designed to answer three interconnected questions. First, is dynamic width expansion better than a fixed-capacity model retaining previously acquired knowledge in the Class-IL setting? Second, does the proposed attention mechanism with persistent key-value memory complement width expansion, or does it provide independent benefit? Third, how do these architectural mechanisms interact with established continual learning strategies - regularization, functional, and replay-based - across benchmarks of varying complexity?

To address these questions, we evaluate four architectural configurations - a fixed-capacity baseline, width expansion alone (WE), attention alone, and their combination (WE + Attention) - across two benchmarks: Split MNIST, a controlled setting using multilayer perceptrons, and Split CIFAR-100, a more demanding visual recognition scenario using a convolutional network. In both cases, the same continual learning strategies (EWC, SI, LwF, LwM, ER, A-GEM) are applied under identical hyperparameter conditions, enabling direct comparison. Lower and upper bounds are provided by sequential training without mitigation and joint training on all data simultaneously, respectively. This structure allows us to isolate the contribution of each architectural component and assess its combined effect across strategy families and task complexities.



\subsection{MNIST Setup}
The split MNIST benchmark \citep{hsu_re-evaluating_2019, van2022three} provides a controlled incremental learning scenario in which the original MNIST dataset is partitioned into five sequential steps, each introducing two new classes. During training, the model receives data exclusively from the current step and must learn to recognize the newly introduced classes while maintaining performance on all previously observed ones. No task identity is provided during inference, following the standard Class-IL protocol.

\begin{figure*}
    \centering
    \includegraphics[width=1\textwidth]{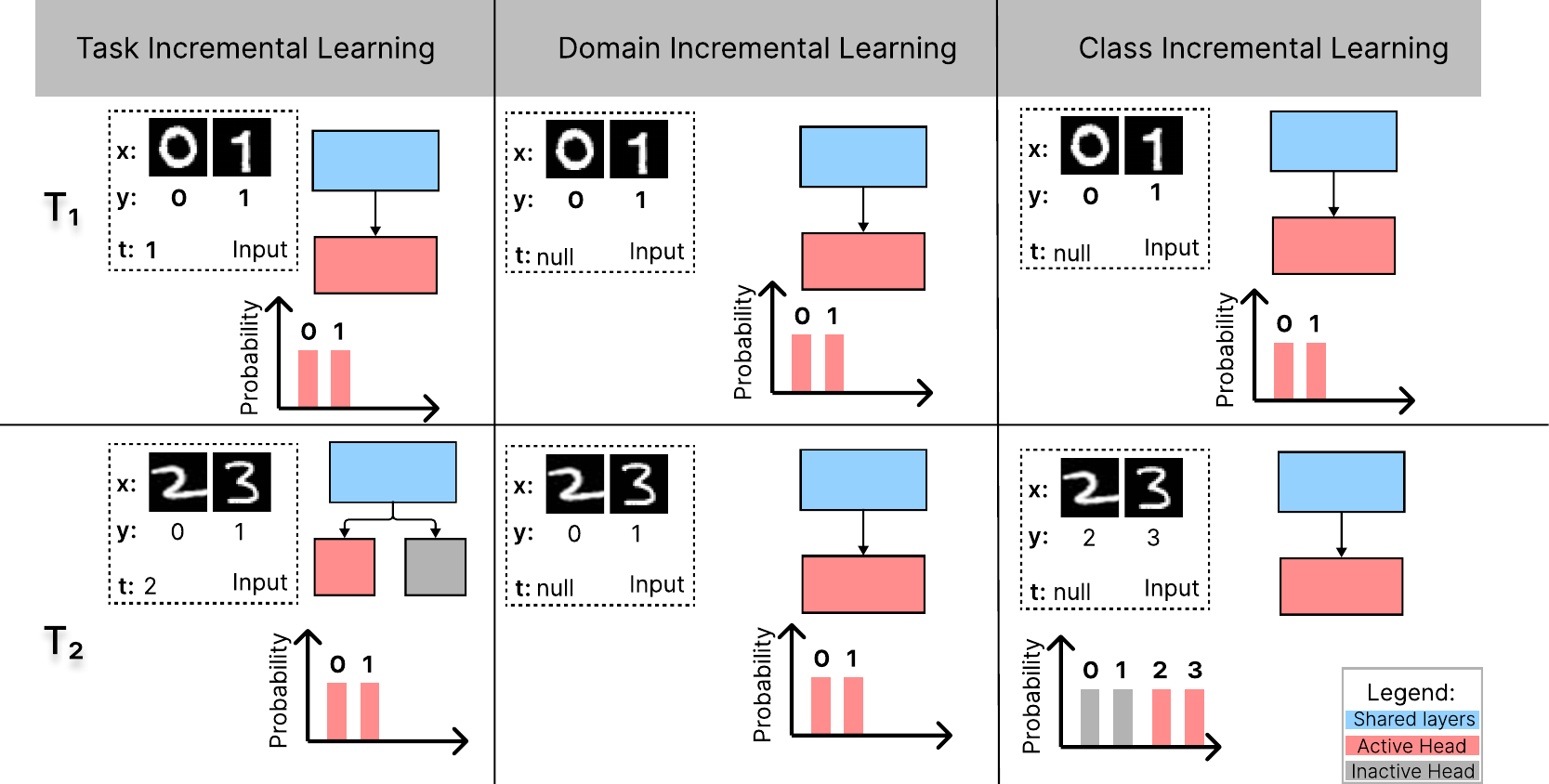}
    \caption{Example of Split MNIST and classification along the three types of incremental learning.}
    \label{fig:SplitMnistDataset}
\end{figure*}
 

All models are trained for five epochs per incremental step using the Adam optimizer with a learning rate of \(0.001\) and momentum parameters \(\beta_1=0.9\) and \(\beta_2=0.999\). Training is performed with mini-batches of 128 samples, while evaluation uses batches of 64 samples. The optimization objective is the cross-entropy loss, computed over the set of classes observed up to the current incremental step. For models incorporating the attention mechanism, the attention dimensionality is set to \(d=128\), and both the key and value memory matrices \(k_{mem}\) and \(v_{mem}\) contain \(32\) vectors each.

Two reference training regimes are considered to bound the expected performance range. The Lower Bound corresponds to sequential training without any forgetting mitigation strategy, representing the worst-case scenario. The Upper Bound corresponds to joint training, an idealized regime in which the model is trained simultaneously on data from all steps, representing the performance ceiling that continual learning methods seek to approximate.

To provide a comprehensive comparison, several widely used continual learning strategies are evaluated under identical experimental conditions. For EWC and SI, the regularization strength is set to \(\lambda=10^9\); while no universally standard value exists for this hyperparameter, values determined empirically via log-scale grid search typically fall between \(10^3\) and \(10^15\) for both datasets in the Class-IL setting \citep{kruengkrai_mitigating_2022,van2022three}. For LwF, the distillation loss uses a temperature parameter \(T=2\) and a weighting coefficient \(\beta=1\), following the common formulation in literature. Replay-based methods maintain an episodic memory buffer of 1000 samples, storing up to 100 samples per class. In the case of A-GEM, the episodic memory also contains 100 samples per class, and reference gradients are computed using mini-batches of 128 samples drawn from the replay buffer.

\subsubsection{Model Architectures}
Four model configurations are evaluated to isolate and analyze the individual and combined effects of width expansion and the proposed attention mechanism.

\begin{figure}
    \centering
    \includegraphics[width=1\columnwidth]{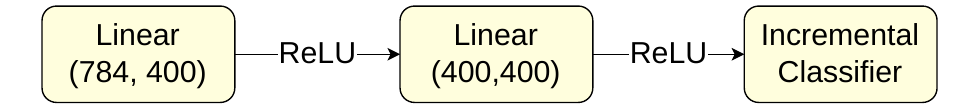}
    \caption{MLP model for MNIST}
    \label{MLP_MNIST}
\end{figure}

\paragraph{Baseline.}
The baseline architecture is a multilayer perceptron with two fully connected hidden layers, each containing 400 neurons, followed by ReLU activations. The network takes a flattened MNIST image as input and produces intermediate feature representations, which are forwarded to a shared incremental classifier that predicts all classes observed up to the current step. The incremental classifier dynamically expands its output dimensionality as new classes are introduced throughout the learning process.
The architecture is illustrated in Figure \ref{MLP_MNIST}.

\begin{figure}
    \centering
    \includegraphics[width=1\columnwidth]{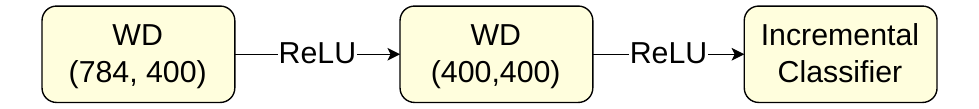}
    \caption{WE model for MNIST}
    \label{WE_MNIST}
\end{figure}

\paragraph{Width Expansion (WE).}
The second configuration introduces the width expansion mechanism described in Section \ref{Expansion_Mechanism}. The standard fully connected layers are replaced with width-dynamic (WD) layers, which are capable of increasing their number of neurons during training based on the normalized loss criterion and neuron utilization statistics. The network is initialized with the same configuration as the baseline model, containing two hidden layers of 400 neurons each. When the expansion criterion is triggered, new neurons are added to the hidden layers while preserving all previously learned weights and connections. The resulting architecture is illustrated in Figure \ref{WE_MNIST}.

\begin{figure*}
    \centering
    \includegraphics[width=0.75\linewidth]{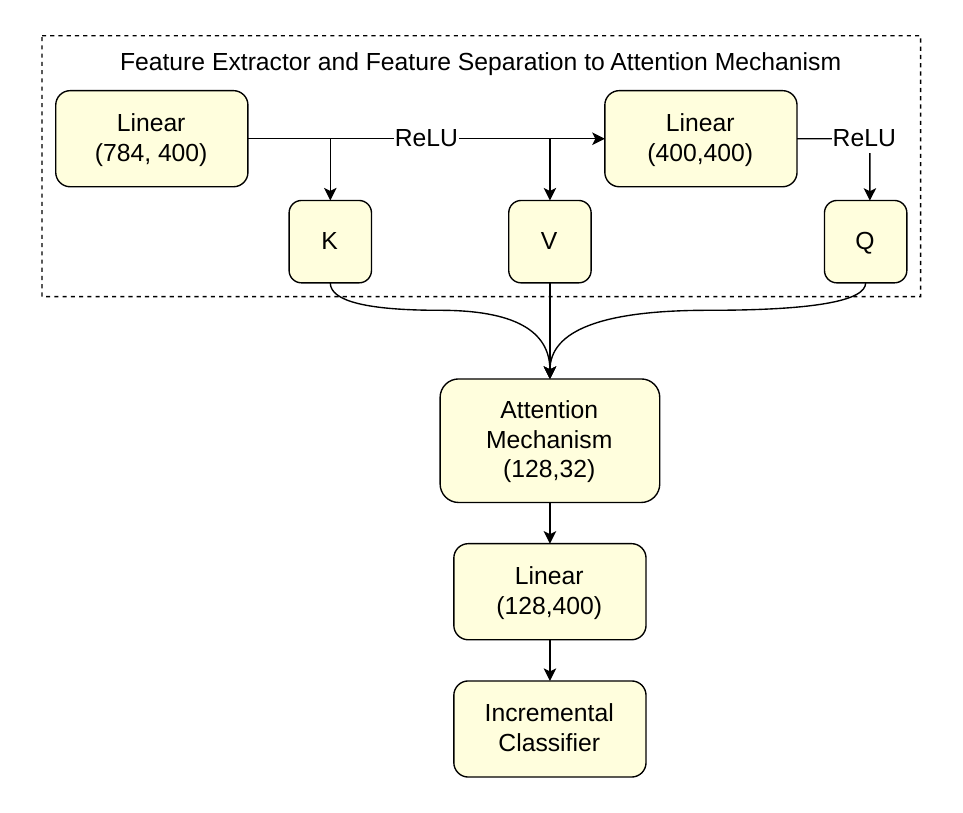}
    \caption{MLP model with attention mechanism for MNIST}
    \label{MLP_Attention_MNIST}
\end{figure*}

\paragraph{MLP with Attention.}
The third configuration augments the baseline architecture with the proposed linear attention module. In this architecture, two fully connected layers are first used to compute intermediate feature representations \(h_1\) and \(h_2\):
\[h_1=ReLU(W_1 x)\]
\[h_2=ReLU(W_2 h_1)\]
The attention module then receives the second-layer representation as the query, while the first-layer representation serves as both key and value:
\[Q=h_2, \;\; K=h_1, \;\; V=h_1\]
Attention is computed as described in Section \ref{Attention_Mechanism}, producing an embedding of dimensionality \(d=128\), which is subsequently projected back into the feature space via an additional fully connected layer before being passed to the incremental classifier. The architecture is illustrated in Figure \ref{MLP_Attention_MNIST}. In this formulation, the attention module enables deeper representations to selectively focus on earlier ones, integrating information from multiple representation levels before generating the final feature vector for classification.

\begin{figure*}
    \centering
    \includegraphics[width=0.75\linewidth]{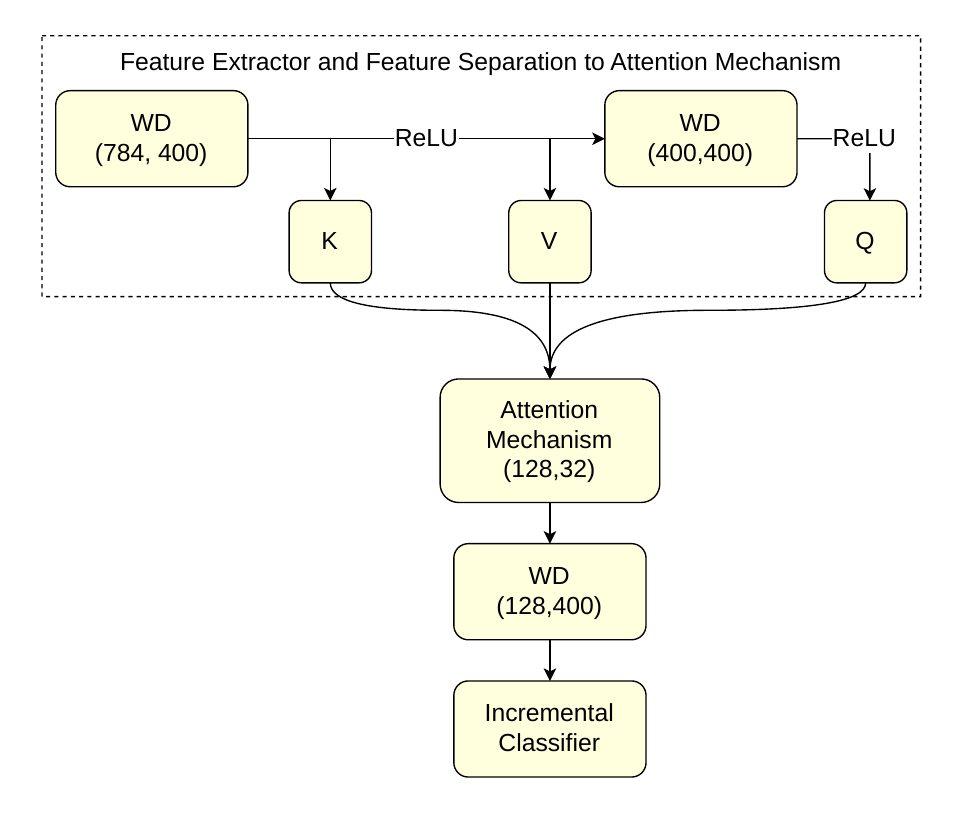}
    \caption{WE model with attention mechanism for MNIST}
    \label{WE_Attention_MNIST}
\end{figure*}

\paragraph{Width Expansion with Attention.}
The fourth configuration combines the width expansion mechanism with the attention module described above. The standard fully connected layers used for feature extraction are replaced with width-dynamic layers, allowing the model to increase its representational capacity during incremental learning as described in Section \ref{Expansion_Mechanism}. The attention module is applied in the same manner as in the previous configuration, with the second-layer serving as query and the first-layer representation as both key and value. The attention output is mapped back into the feature space through a linear projection before being forwarded to the incremental classifier. The architecture is illustrated in Figure \ref{WE_Attention_MNIST}.

This configuration is designed to combine two complementary mechanisms for continual learning: dynamic width expansion increases model capacity as new classes are introduced, while the attention module with persistent memory stabilizes representations across incremental steps, reducing interference between previously and newly learned classes.

\subsection{CIFAR-100 Setup}
The Split CIFAR-100 benchmark \citep{krizhevsky_learning_nodate, iCaRLRebuffi:2017} is used to evaluate the proposed approach in a more challenging visual recognition scenario. The original CIFAR-100 dataset is partitioned into 10 sequential incremental steps, each introducing 10 new classes. As in the Split MNIST protocol, the model receives data exclusively from the current step during training and must maintain performance on all previously observed classes without access to task identity during inference.

\begin{figure}
    \centering
    \includegraphics[width=0.85\columnwidth]{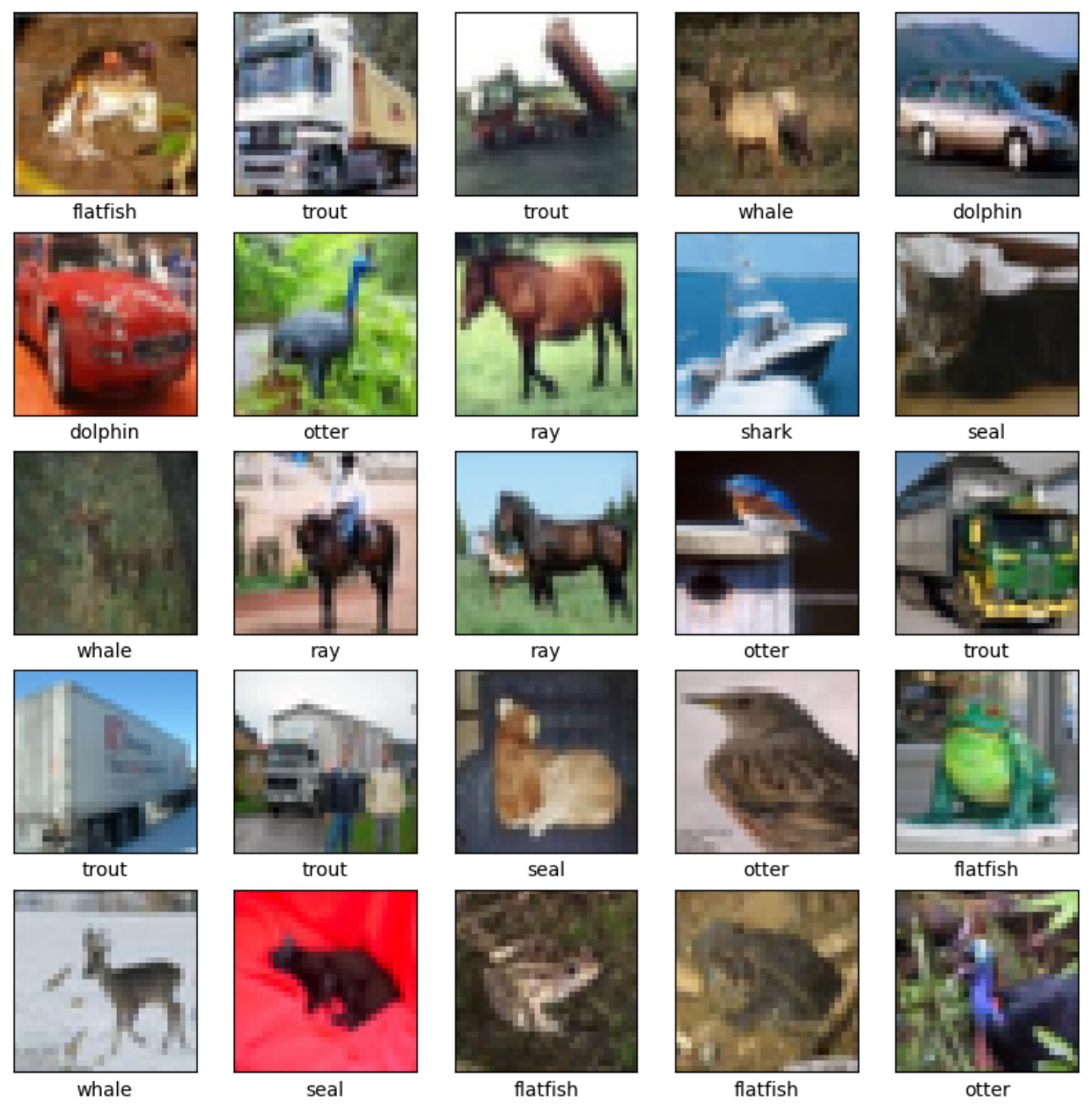}
    \caption{Sample images from the CIFAR-100 dataset illustrating the diversity of object categories and visual appearances.}
    \label{fig:cifar_100_dataset}
\end{figure}

All models are trained using the Adam optimizer with a learning rate of \(0.001\) and momentum parameters \(\beta_1=0.9\) and \(\beta_2=0.999\). Training is performed using mini-batches of 256 samples, while evaluation uses mini-batches of 128 samples. Each incremental step is trained for 30 epochs, except for the upper-bound joint-training configuration, which is trained on the complete dataset for the same number of epochs. The optimization objective is the cross-entropy loss, computed over all classes observed up to the current incremental step. For models that incorporate the attention mechanism, the attention dimensionality is set to \(d=256\), and both the spatial and feature attention memory modules contain \(64\) key-value vectors each.

The same continual learning strategies evaluated on split MNIST are considered here under identical hyperparameter conditions: EWC and SI use a regularization coefficient \(\lambda=10^9\), LwF and LwM use distillation temperature \(T=2\), and weighting coefficient \(\beta=1\). For replay-based methods, an episodic memory with 10,000 samples is maintained with a maximum of 100 samples per class. In the case of A-GEM, the episodic memory also stores 100 samples per class, and reference gradients are computed using batches of 256 samples drawn from the replay buffer.

\subsubsection{Model Architectures}
Four model configurations are evaluated in the convolutional setting to assess the impact of width expansion and the attention mechanism under the greater visual complexity of CIFAR-100.

\begin{figure*}
    \centering
    \includegraphics[width=\linewidth]{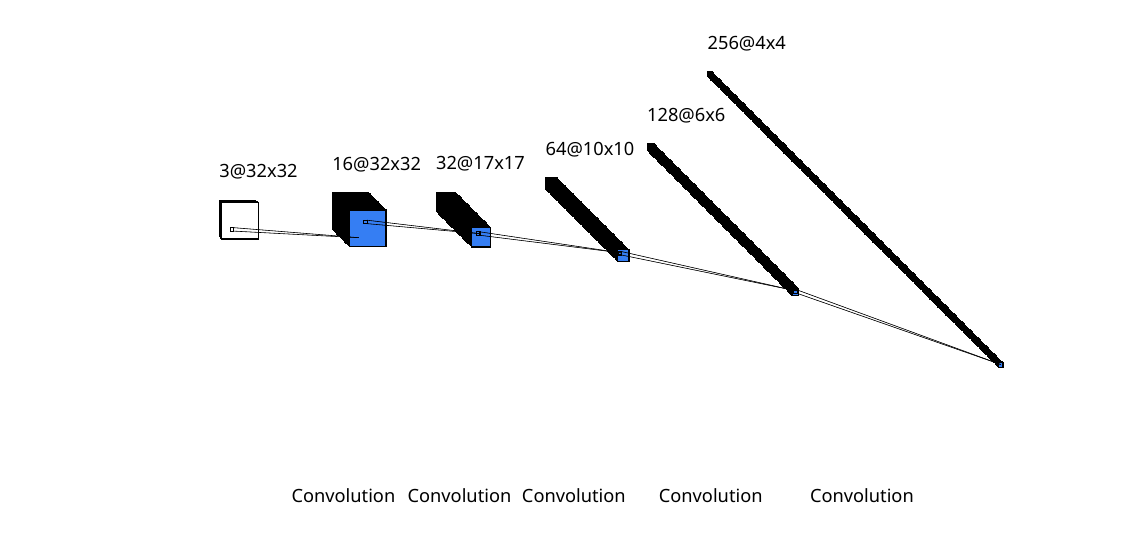}
    \caption{Common feature extractor for CIFAR-100 models}
    \label{backbone}
\end{figure*}

\begin{figure*}
    \centering
    \includegraphics[width=\linewidth]{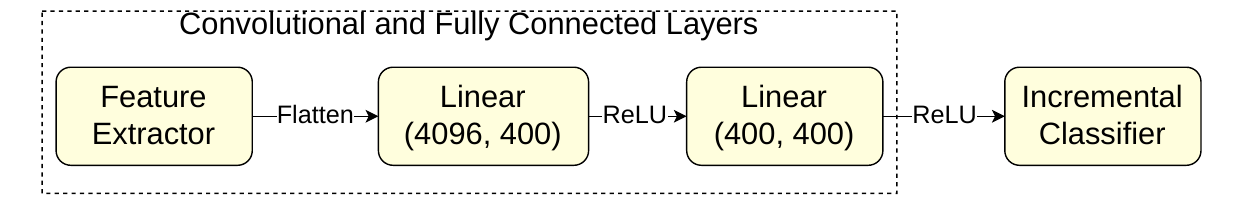}
    \caption{CNN model for CIFAR-100}
    \label{CNN_CIFAR}
\end{figure*}

\paragraph{Baseline.}
The baseline architecture comprises a convolutional feature extractor, two fully connected layers, and a shared incremental classifier. The convolutional backbone consists of five blocks with \(3\times3\) kernels, each followed by batch normalization and ReLU activation. Strided convolutions progressively reduce the spatial resolution from \(32\times32\) to \(4\times4\) while increasing the number of feature channels from \(3\) to \(256\) with the first layer mapping to \(16\) channels and each subsequent doubling that count. The resulting feature map is flattened and processed by two fully connected layers with \(400\) neurons each, producing the final representation used by the incremental classifier. The convolutional backbone is shared across all model configurations evaluated on CIFAR-100, and the full baseline architecture is illustrated in Figures \ref{backbone} and \ref{CNN_CIFAR}.

\begin{figure*}
    \centering
    \includegraphics[width=\linewidth]{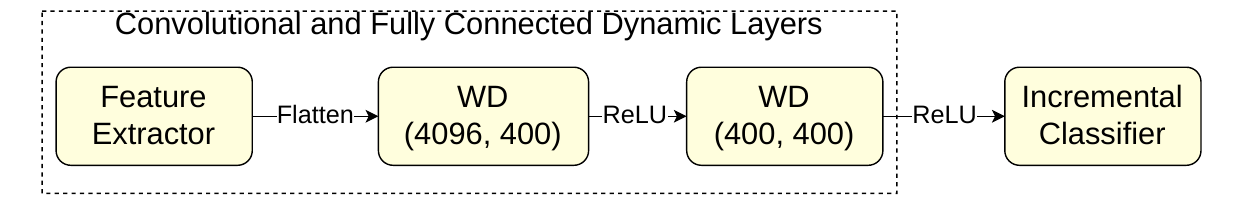}
    \caption{CNN model for CIFAR-100 with width dynamic layers}
    \label{CNNWD_CIFAR}
\end{figure*}

\paragraph{Width Expansion (WE).}
The second configuration introduces the width expansion mechanism in the convolutional setting. The fully connected layers responsible for feature extraction are replaced by width-dynamic layers, which can increase their number of neurons during training based on the normalized loss and neuron utilization statistics. The model is initialized with the same configuration as the baseline, and new neurons are added to the hidden layers when the expansion criterion is triggered, while previously learned weights and connections are preserved. The resulting architecture is illustrated in Figure \ref{CNNWD_CIFAR}.

\paragraph{CNN with Attention.}
The third configuration augments the baseline convolutional architecture with two attention modules. The first operates over the spatial feature maps produced by the final convolutional layer. The feature tensor \(x \in \mathbb{R}^{B \times C \times H \times W}\) is reshaped into a sequence of \(H \cdot W\) spatial tokens of dimensionality \(C\), over which linear self-attention is applied, allowing the model to capture interactions between spatial regions of the image. The resulting attention output is reshaped back to the original spatial format and combined with the input feature map through a residual connection, followed by batch normalization.

\begin{figure*}
    \centering
    \includegraphics[width=\linewidth]{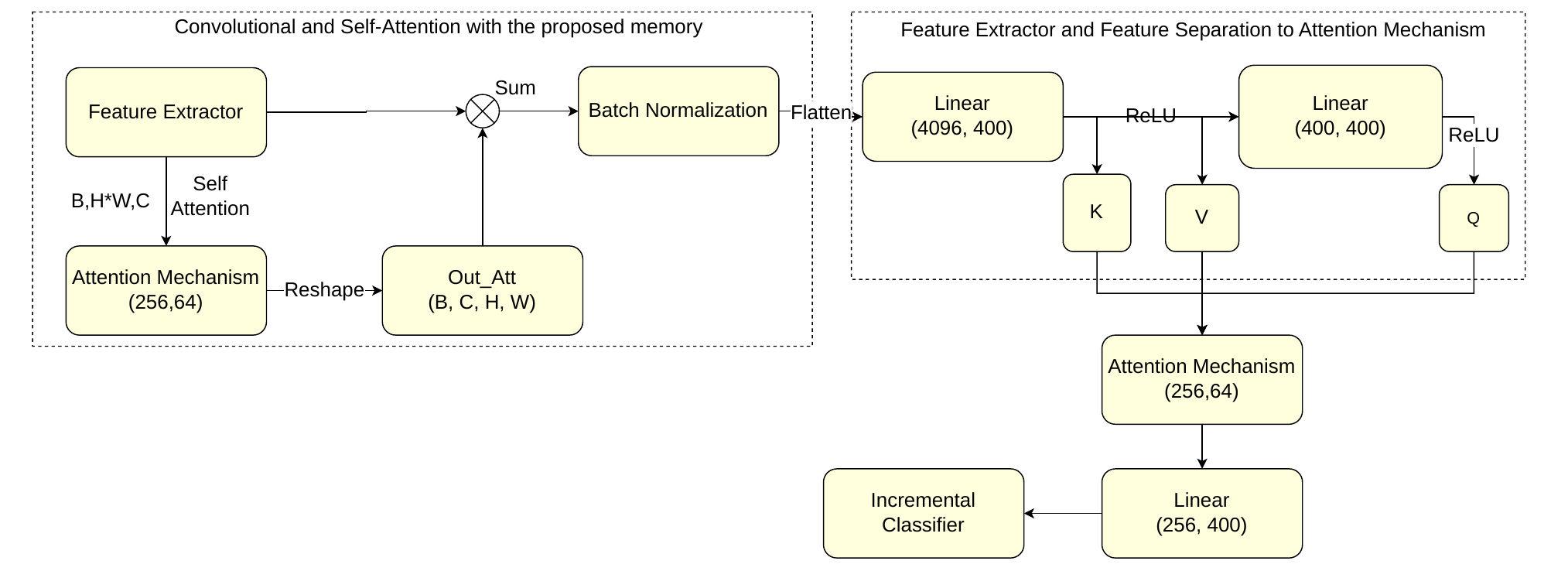}
    \caption{CNN model with attention mechanism for CIFAR-100}
    \label{CNN_Attention_CIFAR}
\end{figure*}

After flattening, two fully connected layers produce intermediate representations \(h_1\) and \(h_2\). A second attention module is then applied in the feature space, following the same cross-level formulation used in the MNIST setting:
\[Q=h_2, \;\;K=h_1, \;\; V=h_1\]
This mechanism allows the deeper representations to selectively attend to early feature representations while leveraging the persistent key-value memory described in Section \ref{Attention_Mechanism}. The attention output is projected back into the feature space through an additional linear layer before being forwarded to the incremental classifier. The full architecture is illustrated in Figure \ref{CNN_Attention_CIFAR}.

\begin{figure*}
    \centering
    \includegraphics[width=\linewidth]{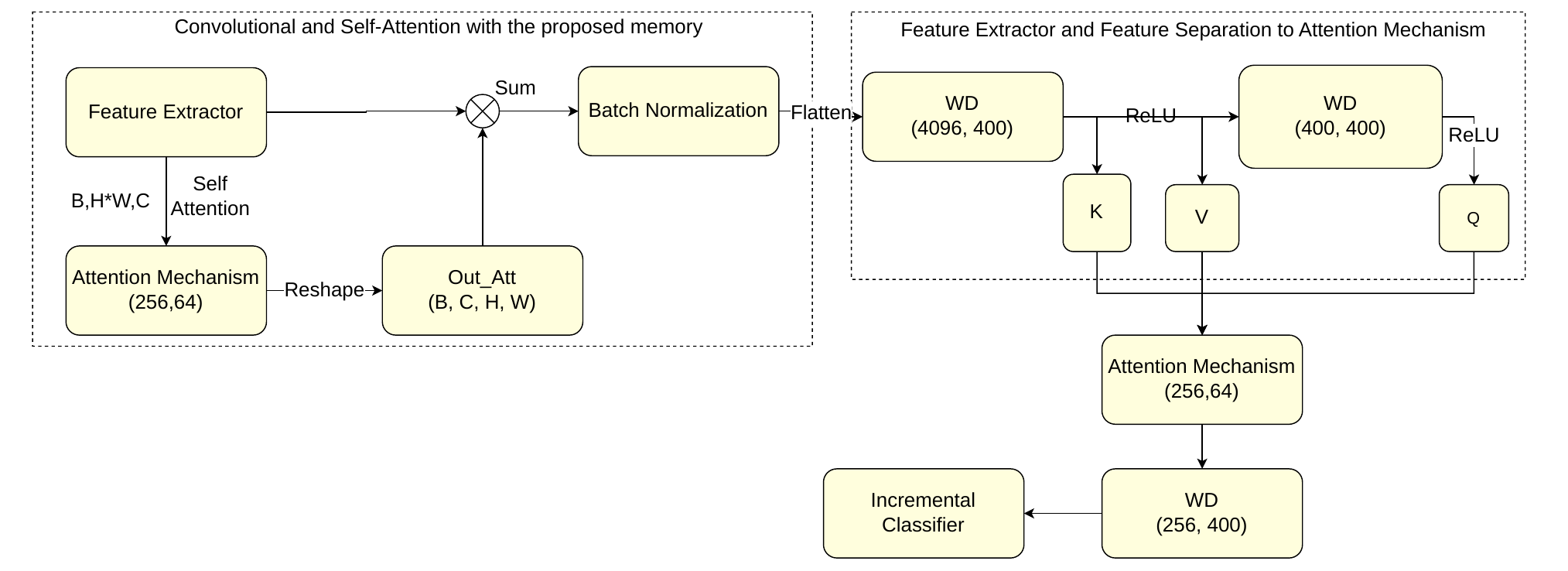}
    \caption{CNNWE model with attention mechanism for CIFAR-100}
    \label{CNNWD_Attention_CIFAR}
\end{figure*}

\paragraph{Width Expansion with Attention.}
The final configuration integrates both the width expansion mechanism and the dual attention modules described above. The fully connected layers used for feature extraction are replaced with width-dynamic layers, allowing the model to increase its representational capacity as new classes are introduced, while both the spatial and feature attention modules are retained. This configuration is designed to jointly exploit two complementary strategies: dynamic capacity growth, which provides additional representational resources when demanded, and attention-based stabilization, which mitigates representational drift and interference across incremental steps. The architecture is illustrated in Figure \ref{CNNWD_Attention_CIFAR}.

\section{Results}\label{Results}
The experimental results reveal clear and consistent patterns across benchmarks, learning strategies, and architectural configurations. Taken together, they highlight the complementary roles of width expansion and the attention mechanism in addressing catastrophic forgetting within the Class-IL setting.

\subsection{Split MNIST}
As shown in Table \ref{results_on_MNIST}, the results on Split MNIST expose a sharp divide between the evaluated strategy families. Replay-based methods, particularly Exemplar Replay, achieve accuracy values close to the upper bound, substantially outperforming all other approaches. In contrast, regularization-based methods - EWC and SI - fail to provide meaningful mitigation of catastrophic forgetting, with results comparable to the lower bound regardless of the architectural configuration employed. This outcome is consistent with prior findings suggesting that parameter-based constraints are poorly suited to the Class-IL setting, where all classes share a common output space and the required plasticity tends to exceed what these methods allow.


\begin{table*}[H]
    \centering
    \begin{tabular}{|c|c|c|c|c|}
         \hline
         \textbf{Strategy} & \textbf{MLP} & \textbf{WE} & \textbf{MLP + Attention} & \textbf{WE + Attention}\\
         \hline
         Joint & \(97.76\) \(\pm0.12\) & - & - & - \\
         \hline
         None & \(19.62\) \(\pm0.07\) & - & - & -\\
         \hline
         EWC & \(19.60\) \(\pm 0.09\) & \(19.41\) \(\pm0.10\) & \(18.67\) \(\pm3.06\) & \(19.75\) \(\pm0.17\) \\
         \hline
         SI & \(19.63\) \(\pm0.05\) & \(19.39\) \(\pm0.11\) & \(19.07\) \(\pm2.59\) & \(19.68\) \(\pm0.04\) \\
         \hline
         LwF & \(29.79\) \(\pm0.85\) & \(29.28\) \(\pm2.00\) & \(39.63\) \(\pm6.25\) & \(44.04\) \(\pm3.93\) \\
         \hline
         ER & \(88.80\) \(\pm0.73\) & \(83.44\) \(\pm0.93\) & \(56.74\) \(\pm33.12\) & \(87.74\) \(\pm0.60\) \\
         \hline
         A-GEM & \(28.34\) \(\pm8.54\) & \(33.97\) \(\pm8.55\) & \(38.38\) \(\pm14.73\) & \(41.25\) \(\pm4.93\) \\
         \hline
    \end{tabular}
    \caption{Results for Split MNIST Values represent Mean \(\pm\) Standard Deviation}
    \label{results_on_MNIST}
\end{table*}

Among functional regularization methods, LwF demonstrates a considerably more favorable response to the proposed architectural modifications. The combination of width expansion and attention achieves the highest accuracy within this group, reaching \(44.04\%\), compared to \(29.79\%\) for the fixed MLP baseline. This improvement reflects the contribution of both mechanisms: width expansion provides additional representational capacity as new classes are introduced, while the attention module with persistent memory stabilizes feature representations across incremental steps, reducing the interference that distillation alone cannot fully prevent. Notably, the attention mechanism isolation introduces grater training variability, as evidenced by the higher standard deviation observed for the MLP with Attention configuration under LwF. When combined with width expansion, however, the model becomes more stable, suggesting that the two mechanisms interact constructively rather than independently.


\begin{figure}
    \centering
    \includegraphics[width=\columnwidth]{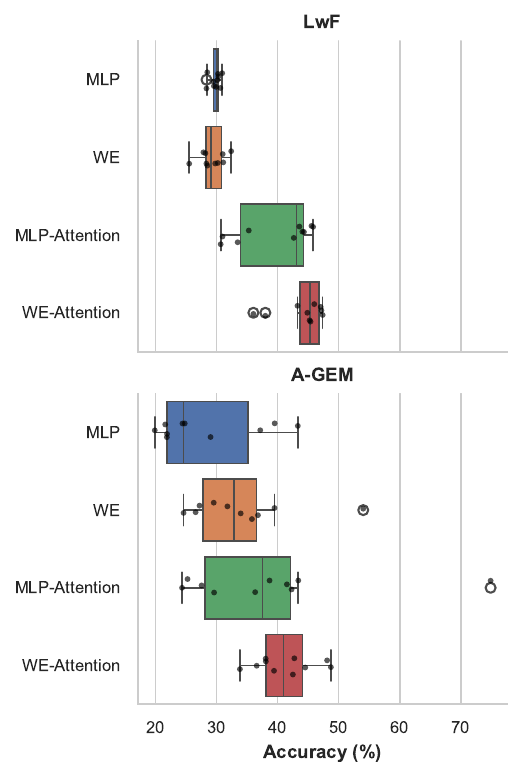}
    \caption{Comparison between LwF and A-GEM on Split-Mnist}
    \label{LwF_AGEM_MNIST}
\end{figure}

A-GEM also benefits consistently from the proposed modifications. Accuracy increases from \(28.34\%\) for the fixed MLP to \(41.25\%\) for the width expansion with attention configuration, with width expansion contributing a meaningful share of this gain even before the attention module is incorporated. As illustrated in Figure \ref{LwF_AGEM_MNIST}, the improvements obtained by architectural enhancements are more pronounced for the LwF than for A-GEM. This asymmetry is interpretable: replay-based methods already mitigate forgetting through explicit sample reuse, and the marginal benefit of representational improvements is therefore smaller than for functional methods, which rely entirely on structural mechanisms to preserve prior knowledge.

It is worth noting that ER exhibits an unexpected drop in performance under the MLP with attention configuration, with a large associated standard deviation. This instability likely to reflects the sensitivity of the attention module to training dynamics in the absence of additional capacity, and is substantially reduced when attention is combined with width expansion.


\subsection{Split CIFAR-100 without Pretraining}
As shown in Table \ref{results_on_CIFAR}, all methods experience a significant reduction in performance on Split CIFAR-100, reflecting the considerably greater visual complexity of this benchmark relative to Split MNIST. Regularization-based approaches again fail to provide meaningful improvements over the lower bound, and this pattern holds across all architectural configurations, reinforcing the conclusion that parameter-based constraints are insufficient to address the challenges of Class-IL in complex visual settings.

\begin{table*}[H]
    \centering
    \begin{tabular}{|c|c|c|c|c|}
         \hline
         \textbf{Strategy} & \textbf{CNN} & \textbf{WE} & \textbf{CNN + Attention} & \textbf{WE + Attention}\\
         \hline
         Joint & \(48.56\) \(\pm0.51\) & - & - & - \\
         \hline
         None & \(8.05\) \(\pm0.17\) & - & - & -\\
         \hline
         EWC & \(6.46\) \(\pm0.34 \) & \(7.27\) \(\pm0.11\) & \(6.49\) \(\pm0.30\) & \(7.53\) \(\pm0.11\) \\
         \hline
         SI & \(5.82\) \(\pm0.16\) & \(8.43\) \(\pm0.08\) & \(5.89\) \(\pm0.17\) & \(8.30\) \(\pm0.09\) \\
         \hline
         LwF & \(12.58\) \(\pm0.30\) & \(14.20\) \(\pm0.34\) & \(15.13\) \(\pm0.56\) & \(17.91\) \(\pm0.90\) \\
         \hline
         LwM & \(10.64\) \(\pm0.26\) & \(12.52\) \(\pm0.38\) & \(10.39\) \(\pm0.85\) & \(12.18\) \(\pm0.86\) \\
         \hline
         ER & \(33.37\) \(\pm0.94\) & \(31.21\) \(\pm0.62\) & \(31.72\) \(\pm0.88\) & \(32.41\) \(\pm0.63\) \\
         \hline
         A-GEM & \(11.72\) \(\pm3.38\) & \(17.91\) \(\pm3.84\) & \(14.39\) \(\pm3.55\) & \(25.17\) \(\pm6.50\) \\
         \hline
    \end{tabular}
    \caption{Results for Split CIFAR-100 without pretraining Values represent Mean \(\pm\) Standard Deviation}
    \label{results_on_CIFAR}
\end{table*}
The proposed width expansion yields consistent improvements across all functional and replay-adjacent strategies. The effect is particularly pronounced for A-GEM, where accuracy increases from \(11.72\%\) to \(17.91\%\) with width expansion alone, and reaches \(25.17\%\) when the attention mechanism is added - the best result among all methods that not rely on direct sample replay. This progression suggests that the two mechanisms provide complementary benefits: expansion increases the representational resources available for new classes, while the attention module mitigates the drift in previously learned representations that expansion alone cannot prevent.

\begin{figure}
    \centering
    \includegraphics[width=\columnwidth]{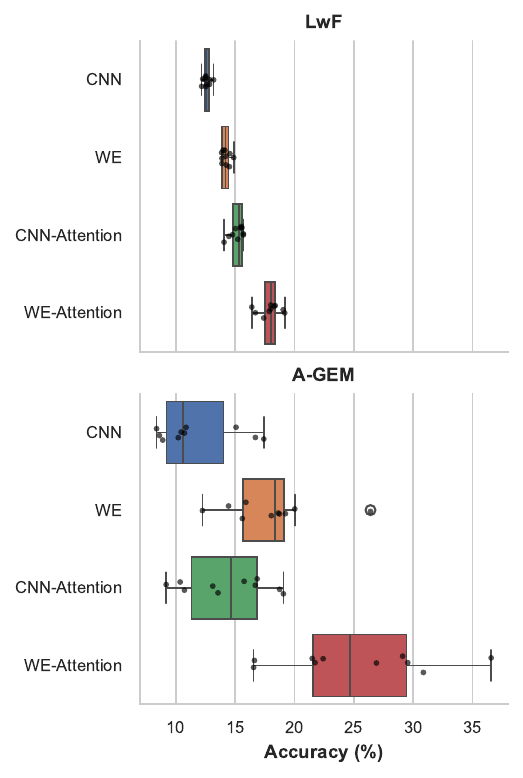}
    \caption{Comparison between LwF and A-GEM on Split CIFAR-100}
    \label{LwF_AGEM_CIFAR100}
\end{figure}

For LwF, width expansion also yields a consistent improvement, with accuracy increasing from \(12.58\%\) to \(14.20\%\) under expansion alone, and to \(17.91\%\) when attention is incorporated. As illustrated in Figure \ref{LwF_AGEM_CIFAR100}, the attention mechanism has a stronger impact when combined with width expansion than when applied in isolation, mirroring the pattern observed on Split MNIST and further supporting the interpretation that attention-based stabilization is most effective when sufficient representational capacity is available.

Among replay-based methods, ER shows no meaningful benefit from width expansion or attention, with results remaining stable across all architectural configurations. This is consistent with the earlier observation that methods relying on stored samples are less sensitive to representational improvements, since explicit rehearsal already provides a strong mechanism for preserving prior knowledge.


\subsection{Split CIFAR-100 with Pretraining}
Table \ref{results_on_CIFAR_With_pretraining} reports results for the convolutional setting in which the backbone is initialized with weights pretrained on CIFAR-10 and subsequently frozen during incremental training. Although pretraining might intuitively be expected to provide a stronger representational foundation, the results show that models trained from scratch outperform their pretrained counterparts across nearly all strategies and configurations.


\begin{table*}[H]
    \centering
    \begin{tabular}{|c|c|c|c|c|}
         \hline
         \textbf{Strategy} & \textbf{CNN} & \textbf{WE} & \textbf{CNN + Attention} & \textbf{WE + Attention}\\
         \hline
         Joint & \(44.09\) \(\pm0.40\) & - & - & - \\
         \hline
         None & \(8.04\) \(\pm0.06\) & - & - & -\\
         \hline
         EWC & \(7.58\) \(\pm0.24 \) & \(7.64\) \(\pm0.17\) & \(7.83\) \(\pm0.14\) & \(7.86\) \(\pm0.15\) \\
         \hline
         SI & \(7.35\) \(\pm0.11\) & \(7.38\) \(\pm0.06\) & \(5.61\) \(\pm0.21\) & \(5.65\) \(\pm0.22\) \\
         \hline
         LwF & \(14.19\) \(\pm0.32\) & \(14.43\) \(\pm0.40\) & \(16.13\) \(\pm0.59\) & \(16.41\) \(\pm0.69\) \\
         \hline
         ER & \(31.72\) \(\pm0.26\) & \(31.60\) \(\pm0.56\) & \(33.38\) \(\pm0.57\) & \(33.28\) \(\pm0.38\) \\
         \hline
         A-GEM & \(8.10\) \(\pm0.14\) & \(8.15\) \(\pm0.10\) & \(13.23\) \(\pm1.67\) & \(15.37\) \(\pm4.94\) \\
         \hline
    \end{tabular}
    \caption{Results for Split CIFAR-100 with pretraining values represent Mean \(\pm\) Standard Deviation}
    \label{results_on_CIFAR_With_pretraining}
\end{table*}

This counterintuitive outcome can be attributed to the representational mismatch between CIFAR-10 and CIFAR-100. Because the convolutional layers are frozen after pretraining, the model is constrained to a fixed feature space that was optimized for a simpler and categorically different distribution. The resulting representations lack the expressiveness required for the finer-grained distinctions among CIFAR-100 classes, forcing the classifier and the proposed expansion mechanisms to compensate for suboptimal features - a task that fundamentally limits achievable performance.

Pretraining does, however, reduce training variability: pretrained models exhibit smaller standard deviations across runs, acting as a strong implicit regularizer. This stability comes at a clear cost to plasticity, as frozen convolutional layers cannot adapt their representations to the new data distribution encountered during incremental learning. The proposed width expansion, which operates exclusively on the fully connected layers in this configuration, can only partially compensate for the rigidity of the fixed feature extractor.



\begin{figure}
    \centering
    \includegraphics[width=\columnwidth]{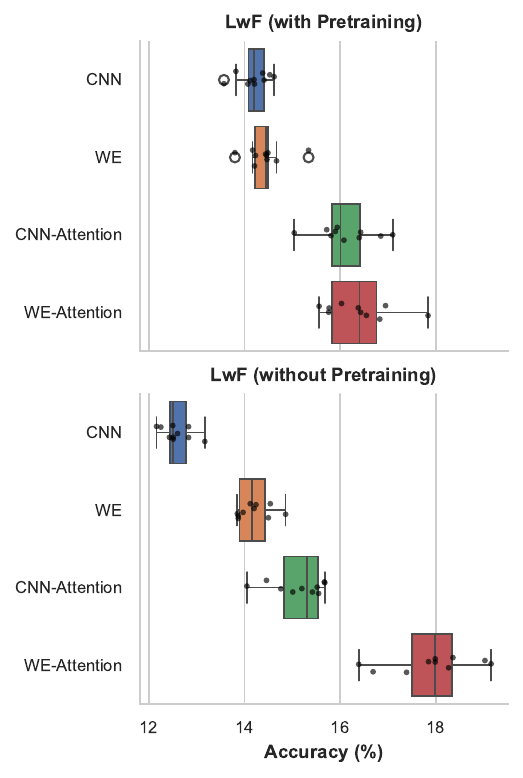}
    \caption{LwF comparison between with and without pretraining.}
    \label{LwF_with_without_pretraining}
\end{figure}

\begin{figure}
    \centering
    \includegraphics[width=\columnwidth]{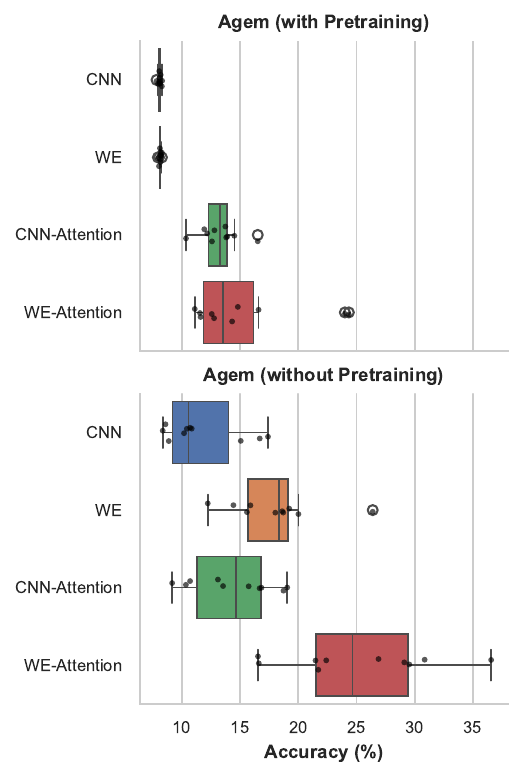}
    \caption{A-GEM comparison between with and without pretraining.}
    \label{AGEM_with_without_pretraining}
\end{figure}

As illustrated in Figures \ref{LwF_with_without_pretraining} and \ref{AGEM_with_without_pretraining}, pretrained models consistently underperform their non-pretrained counterparts across both LwF and A-GEM, with the gap particularly evident in the A-GEM setting. This pattern reinforces a central conclusion of the study: in the Class-IL setting, the ability to adapt feature representations incrementally is more important than the quality of the initial representation space. A well-initialized but fixed feature extractor introduces constraints that ultimately hinder the kind of flexible adaptation that continual learning demands.



\section{Threats to Validity}\label{Threats}
A critical assessment of this study's methodology and results reveals several potential threats to the validity of its conclusions. Acknowledging these limitations is essential for contextualizing the findings and guiding future research.

\subsection{Internal Validity}
Internal validity concerns whether the observed improvements can be attributed to the proposed mechanisms rather than confounding factors. Several design choices support this claim. Hyperparameters for the baseline strategies (\(\lambda=10^9\) for EWC and SI, \(T=2\) and \(\beta=1\) for LwF and LwM, buffer sizes of 100 samples per class) were set following established practice in the Class-IL literature \citep{van2022three, kruengkrai_mitigating_2022} and held constant across all architectural configurations, ensuring that strategy-level differences do not inflate observed gains. All results are reported as means and standard deviations across multiple independent runs with different random seeds, providing a measure of result stability. However, the expansion-specific hyperparameters - including loss threshold, growth factor, and weighting coefficients \(w_{loss}\) and \(w_{local}\) - were not subjected to systematic held-out grid search, and their interaction with specific strategy-architecture combinations cannot be fully ruled out as a source of variance. The high standard deviation observed for A-GEM under some configurations (e.g., 14.73 on Split MNIST under MLP + Attention) suggests that certain combinations are sensitive to initialization or training dynamics in ways not fully captured by the reported statistics.

\subsection{External Validity}
External validity concerns the generalization of findings beyond the evaluated conditions. Two benchmarks of substantially different complexity were used: Split MNIST, a controlled setting, and Split CIFAR-100, a more demanding visual recognition task. This range provides meaningful coverage of the Class-IL landscape. Nevertheless, both benchmarks use balanced class splits of equal and fixed size, a condition that may not hold in real-world deployments, where class arrival frequency, distributional shift, and task duration may vary unpredictably. The convolutional backbone used for CIFAR-100 is relatively shallow compared to architectures used in recent state-of-the-art work; performance on large-scale benchmarks such as Split ImageNet, or in settings based on pre-trained transformer backbones, cannot be directly inferred from the present results. Additionally, all evaluations assume a fixed incremental schedule with clearly delineated steps; behavior of the expansion mechanism under irregular, fine-grained, or blurred task boundaries remains an open question.

\subsection{Construct Validity}
Construct validity concerns whether the evaluation metrics adequately capture the properties of interest. Average classification across all observed classes at the end of training is used as the primary metric, consistent with the standard Class-IL evaluation protocol \citep{van2022three}. This metric, however, does not distinguish between forgetting of early classes and failure to learn later ones - two failure modes with distinct implications for the stability-plasticity trade-off. Furthermore, it does not capture the computational cost incurred by dynamic expansion, which is a practically relevant consideration in resource-constrained deployment scenarios. Future evaluations should complement accuracy with backward transfer, forward transfer, and parameter count over time to provide a more complete characterization of each mechanism's contribution.

\section{Conclusion}\label{Conclusion}
This work investigated catastrophic forgetting in Class-IL through the lens of architectural adaptability, proposing a method based on dynamic width expansion within existing network layers, complemented by a linear attention mechanism with a persistent key-value memory. Unlike network expansion approaches such as DER and DNE, which grow model capacity through the addition of task-specific modules and therefore depend on explicit task identifiers, the proposed method expands representational capacity within existing layers based on a normalized cross-entropy criterion that reflects representational demand. This design makes the approach directly applicable to the Class-IL setting, where task boundaries are not available during inference.

Experimental results across Split MNIST and Split CIFAR-100 demonstrate that the combination of width expansion and the attention mechanism provides consistent improvements in knowledge retention. On Split MNIST, the width expansion with the attention module configuration achieves the highest accuracy among functional regularization methods, reaching \(44.04\%\), while on Split CIFAR-100, the same configuration yields \(25.17\%\) - the best result among methods that do not rely on direct sample replay. Across both benchmarks, the attention mechanism with persistent key-value memory plays a central role in stabilizing feature representations and mitigating representational drift during incremental updates. The results also reveal that the two mechanisms interact constructively: attention-based stabilization is most effective when combined with sufficient representational capacity, and width expansion alone yields more variable results in the absence of the representational anchoring provided by the persistent memory.

A recurring and significant finding across all experiments is that adaptability in feature extraction layers proves more consequential than the quality of the initial representation space. Models trained from scratch consistently outperformed those with pretrained and frozen convolutional backbones, as the latter introduce representational constraints that limit the plasticity required to accommodate new data distributions over time. This result suggests that, in complex continual learning scenarios, the capacity to revise learned representations incrementally should be prioritized over initialization-based stability.

Several directions emerge from the present work. First, the expansion mechanism currently operates on fully connected layers; extending it to convolutional layers could allow the model to adapt its feature extraction capacity directly, rather than relying solely on downstream dense layers to compensate for representational bottlenecks. Second, selective post-expansion pruning could recover computational efficiency without sacrificing accuracy, addressing the risk of unbounded parameter growth observed in the current results. Third, the normalized global loss criterion used to trigger expansion could be replaced by layer-wise or neuron-wise importance signals, enabling finer-grained allocation of representational resources across the network. Finally, evaluation on larger-scale and class-imbalanced benchmarks, as well as in settings with irregular overlapping task boundaries, would provide a more complete picture of the approach's applicability to real-world continual learning scenarios.

\printcredits

\section*{Acknowledgements}
This work was partially supported by the São Paulo Research Foundation (FAPESP), grant 2025/13241-3. The Article Processing Charge (APC) was funded by the Brazilian Federal Agency for Support and Evaluation of Graduate Education -- CAPES (ROR identifier: 00x0ma614). For the purposes of open access, the authors have applied a Creative Commons CC BY license to any accepted version of the article.

\bibliographystyle{cas-model2-names}

\bibliography{cas-refs}

@InProceedings{LinearAttetionKatharopoulos:2020,
  title = 	 {Transformers are {RNN}s: Fast Autoregressive Transformers with Linear Attention},
  author =       {Katharopoulos, Angelos and Vyas, Apoorv and Pappas, Nikolaos and Fleuret, Fran{\c{c}}ois},
  booktitle = 	 {Proceedings of the 37th International Conference on Machine Learning},
  pages = 	 {5156--5165},
  year = 	 {2020},
  editor = 	 {III, Hal Daumé and Singh, Aarti},
  volume = 	 {119},
  series = 	 {Proceedings of Machine Learning Research},
  month = 	 {13--18 Jul},
  publisher =    {PMLR},
  url = 	 {https://proceedings.mlr.press/v119/katharopoulos20a.html}
}

@inproceedings{kim2023stability,
  title={On the stability-plasticity dilemma of class-incremental learning},
  author={Kim, Dongwan and Han, Bohyung},
  booktitle={Proceedings of the IEEE/CVF conference on computer vision and pattern recognition},
  pages={20196--20204},
  year={2023}
}

@article{cao2022beyond,
  title={Beyond iid: Non-iid thinking, informatics, and learning},
  author={Cao, Longbing},
  journal={IEEE Intelligent Systems},
  volume={37},
  number={4},
  pages={5--17},
  year={2022},
  publisher={IEEE}
}

@article{EWCKirkpatrick:2017,
author = {James Kirkpatrick  and Razvan Pascanu  and Neil Rabinowitz  and Joel Veness  and Guillaume Desjardins  and Andrei A. Rusu  and Kieran Milan  and John Quan  and Tiago Ramalho  and Agnieszka Grabska-Barwinska  and Demis Hassabis  and Claudia Clopath  and Dharshan Kumaran  and Raia Hadsell },
title = {Overcoming catastrophic forgetting in neural networks},
journal = {Proceedings of the National Academy of Sciences},
volume = {114},
number = {13},
pages = {3521-3526},
year = {2017},
doi = {10.1073/pnas.1611835114},
URL = {https://www.pnas.org/doi/abs/10.1073/pnas.1611835114},
eprint = {https://www.pnas.org/doi/pdf/10.1073/pnas.1611835114}}

@ARTICLE{SIZenke:2017,
  title    = "Continual Learning Through Synaptic Intelligence",
  author   = "Zenke, Friedemann and Poole, Ben and Ganguli, Surya",
  journal  = "Proc Mach Learn Res",
  volume   =  70,
  pages    = "3987--3995",
  year     =  2017,
  address  = "United States",
  language = "en"
}

@article{LwFLi:2017,
  title={Learning without forgetting},
  author={Li, Zhizhong and Hoiem, Derek},
  journal={IEEE transactions on pattern analysis and machine intelligence},
  volume={40},
  number={12},
  pages={2935--2947},
  year={2017},
  publisher={IEEE}
}

@inproceedings{iCaRLRebuffi:2017,
  title={icarl: Incremental classifier and representation learning},
  author={Rebuffi, Sylvestre-Alvise and Kolesnikov, Alexander and Sperl, Georg and Lampert, Christoph H},
  booktitle={Proceedings of the IEEE conference on Computer Vision and Pattern Recognition},
  pages={2001--2010},
  year={2017}
}

@inproceedings{LwMDhar:2019,
  title={Learning without memorizing},
  author={Dhar, Prithviraj and Singh, Rajat Vikram and Peng, Kuan-Chuan and Wu, Ziyan and Chellappa, Rama},
  booktitle={Proceedings of the IEEE/CVF conference on computer vision and pattern recognition},
  pages={5138--5146},
  year={2019}
}

@inproceedings{
AGEMChaudhry:2019,
title={Efficient Lifelong Learning with A-{GEM}},
author={Arslan Chaudhry and Marc’Aurelio Ranzato and Marcus Rohrbach and Mohamed Elhoseiny},
booktitle={International Conference on Learning Representations},
year={2019},
url={https://openreview.net/forum?id=Hkf2_sC5FX},
}

@INPROCEEDINGS{DERYan:2021,
  author={Yan, Shipeng and Xie, Jiangwei and He, Xuming},
  booktitle={2021 IEEE/CVF Conference on Computer Vision and Pattern Recognition (CVPR)}, 
  title={DER: Dynamically Expandable Representation for Class Incremental Learning}, 
  year={2021},
  pages={3013-3022},
  doi={10.1109/CVPR46437.2021.00303}}

@InProceedings{DNEHu:2023,
    author    = {Hu, Zhiyuan and Li, Yunsheng and Lyu, Jiancheng and Gao, Dashan and Vasconcelos, Nuno},
    title     = {Dense Network Expansion for Class Incremental Learning},
    booktitle = {Proceedings of the IEEE/CVF Conference on Computer Vision and Pattern Recognition (CVPR)},
    month     = {June},
    year      = {2023},
    pages     = {11858-11867}
}

@article{CLDNNRao:2020,
  title={Embracing change: Continual learning in deep neural networks},
  author={Hadsell, Raia and Rao, Dushyant and Rusu, Andrei A and Pascanu, Razvan},
  journal={Trends in cognitive sciences},
  volume={24},
  number={12},
  pages={1028--1040},
  year={2020},
  publisher={Elsevier}
}

@article{van2022three,
  title={Three types of incremental learning},
  author={Van de Ven, Gido M and Tuytelaars, Tinne and Tolias, Andreas S},
  journal={Nature Machine Intelligence},
  volume={4},
  number={12},
  pages={1185--1197},
  year={2022},
  publisher={Nature Publishing Group UK London}
}

@inproceedings{kruengkrai_mitigating_2022,
    address = {Gyeongju, Republic of Korea},
    title = {Mitigating the {Diminishing} {Effect} of {Elastic} {Weight} {Consolidation}},
    url = {https://aclanthology.org/2022.coling-1.403/},
    urldate = {2026-04-15},
    booktitle = {Proceedings of the 29th {International} {Conference} on {Computational} {Linguistics}},
    publisher = {International Committee on Computational Linguistics},
    author = {Kruengkrai, Canasai and Yamagishi, Junichi},
    editor = {Calzolari, Nicoletta and Huang, Chu-Ren and Kim, Hansaem and Pustejovsky, James and Wanner, Leo and Choi, Key-Sun and Ryu, Pum-Mo and Chen, Hsin-Hsi and Donatelli, Lucia and Ji, Heng and Kurohashi, Sadao and Paggio, Patrizia and Xue, Nianwen and Kim, Seokhwan and Hahm, Younggyun and He, Zhong and Lee, Tony Kyungil and Santus, Enrico and Bond, Francis and Na, Seung-Hoon},
    month = oct,
    year = {2022},
    pages = {4568--4574},
}

@inproceedings{ER:2019,
 author = {Rolnick, David and Ahuja, Arun and Schwarz, Jonathan and Lillicrap, Timothy and Wayne, Gregory},
 booktitle = {Advances in Neural Information Processing Systems},
 editor = {H. Wallach and H. Larochelle and A. Beygelzimer and F. d\textquotesingle Alch\'{e}-Buc and E. Fox and R. Garnett},
 pages = {},
 publisher = {Curran Associates, Inc.},
 title = {Experience Replay for Continual Learning},
 url = {https://proceedings.neurips.cc/paper_files/paper/2019/file/fa7cdfad1a5aaf8370ebeda47a1ff1c3-Paper.pdf},
 volume = {32},
 year = {2019}
}

@ARTICLE{CILSurvey:2024,
  author={Zhou, Da-Wei and Wang, Qi-Wei and Qi, Zhi-Hong and Ye, Han-Jia and Zhan, De-Chuan and Liu, Ziwei},
  journal={IEEE Transactions on Pattern Analysis and Machine Intelligence}, 
  title={Class-Incremental Learning: A Survey}, 
  year={2024},
  volume={46},
  number={12},
  pages={9851-9873},
  doi={10.1109/TPAMI.2024.3429383}}

@inproceedings{zhou_forward_2022,
    address = {New Orleans, LA, USA},
    title = {Forward {Compatible} {Few}-{Shot} {Class}-{Incremental} {Learning}},
    copyright = {https://doi.org/10.15223/policy-029},
    isbn = {978-1-6654-6946-3},
    url = {https://ieeexplore.ieee.org/document/9878986/},
    doi = {10.1109/CVPR52688.2022.00884},
    language = {en},
    urldate = {2026-05-04},
    booktitle = {2022 {IEEE}/{CVF} {Conference} on {Computer} {Vision} and {Pattern} {Recognition} ({CVPR})},
    publisher = {IEEE},
    author = {Zhou, Da-Wei and Wang, Fu-Yun and Ye, Han-Jia and Ma, Liang and Pu, Shiliang and Zhan, De-Chuan},
    month = jun,
    year = {2022},
    pages = {9036--9046},
}

@inproceedings{de_lange_continual_2021,
    address = {Montreal, QC, Canada},
    title = {Continual {Prototype} {Evolution}: {Learning} {Online} from {Non}-{Stationary} {Data} {Streams}},
    copyright = {https://doi.org/10.15223/policy-029},
    isbn = {978-1-6654-2812-5},
    shorttitle = {Continual {Prototype} {Evolution}},
    url = {https://ieeexplore.ieee.org/document/9711397/},
    doi = {10.1109/ICCV48922.2021.00814},
    language = {en},
    urldate = {2026-05-04},
    booktitle = {2021 {IEEE}/{CVF} {International} {Conference} on {Computer} {Vision} ({ICCV})},
    publisher = {IEEE},
    author = {De Lange, Matthias and Tuytelaars, Tinne},
    month = oct,
    year = {2021},
    pages = {8230--8239},
}

@inproceedings{zhao_maintaining_2020,
    title = {Maintaining {Discrimination} and {Fairness} in {Class} {Incremental} {Learning}},
    issn = {2575-7075},
    url = {https://ieeexplore.ieee.org/document/9156766/},
    doi = {10.1109/CVPR42600.2020.01322},
    urldate = {2026-05-05},
    booktitle = {2020 {IEEE}/{CVF} {Conference} on {Computer} {Vision} and {Pattern} {Recognition} ({CVPR})},
    author = {Zhao, Bowen and Xiao, Xi and Gan, Guojun and Zhang, Bin and Xia, Shu-Tao},
    month = jun,
    year = {2020},
    note = {ISSN: 2575-7075},
    pages = {13205--13214},
}

@article{liu_incremental_2023,
    title = {Incremental learning with neural networks for computer vision: a survey},
    volume = {56},
    issn = {1573-7462},
    shorttitle = {Incremental learning with neural networks for computer vision},
    url = {https://doi.org/10.1007/s10462-022-10294-2},
    doi = {10.1007/s10462-022-10294-2},
    language = {en},
    number = {5},
    urldate = {2026-05-05},
    journal = {Artificial Intelligence Review},
    author = {Liu, Hao and Zhou, Yong and Liu, Bing and Zhao, Jiaqi and Yao, Rui and Shao, Zhiwen},
    month = may,
    year = {2023},
    pages = {4557--4589},
}

@inproceedings{zhou_expandable_2024,
    address = {Seattle, WA, USA},
    title = {Expandable {Subspace} {Ensemble} for {Pre}-{Trained} {Model}-{Based} {Class}-{Incremental} {Learning}},
    copyright = {https://doi.org/10.15223/policy-029},
    isbn = {979-8-3503-5300-6},
    url = {https://ieeexplore.ieee.org/document/10656913/},
    doi = {10.1109/CVPR52733.2024.02223},
    language = {en},
    urldate = {2026-05-05},
    booktitle = {2024 {IEEE}/{CVF} {Conference} on {Computer} {Vision} and {Pattern} {Recognition} ({CVPR})},
    publisher = {IEEE},
    author = {Zhou, Da-Wei and Sun, Hai-Long and Ye, Han-Jia and Zhan, De-Chuan},
    month = jun,
    year = {2024},
    pages = {23554--23564},
}

@article{fu_knowledge_2023,
    title = {Knowledge aggregation networks for class incremental learning},
    volume = {137},
    issn = {00313203},
    url = {https://linkinghub.elsevier.com/retrieve/pii/S0031320323000110},
    doi = {10.1016/j.patcog.2023.109310},
    language = {en},
    urldate = {2026-05-05},
    journal = {Pattern Recognition},
    author = {Fu, Zhiling and Wang, Zhe and Xu, Xinlei and Li, Dongdong and Yang, Hai},
    month = may,
    year = {2023},
    pages = {109310},
}

@inproceedings{yu_semantic_2020,
    address = {Seattle, WA, USA},
    title = {Semantic {Drift} {Compensation} for {Class}-{Incremental} {Learning}},
    copyright = {https://ieeexplore.ieee.org/Xplorehelp/downloads/license-information/IEEE.html},
    isbn = {978-1-7281-7168-5},
    url = {https://ieeexplore.ieee.org/document/9156964/},
    doi = {10.1109/CVPR42600.2020.00701},
    language = {en},
    urldate = {2026-05-04},
    booktitle = {2020 {IEEE}/{CVF} {Conference} on {Computer} {Vision} and {Pattern} {Recognition} ({CVPR})},
    publisher = {IEEE},
    author = {Yu, Lu and Twardowski, Bartlomiej and Liu, Xialei and Herranz, Luis and Wang, Kai and Cheng, Yongmei and Jui, Shangling and Van De Weijer, Joost},
    month = jun,
    year = {2020},
    pages = {6980--6989},
}

@misc{hsu_re-evaluating_2019,
    title = {Re-evaluating {Continual} {Learning} {Scenarios}: {A} {Categorization} and {Case} for {Strong} {Baselines}},
    shorttitle = {Re-evaluating {Continual} {Learning} {Scenarios}},
    url = {http://arxiv.org/abs/1810.12488},
    doi = {10.48550/arXiv.1810.12488},
    urldate = {2026-05-06},
    publisher = {arXiv},
    author = {Hsu, Yen-Chang and Liu, Yen-Cheng and Ramasamy, Anita and Kira, Zsolt},
    month = jan,
    year = {2019},
    note = {arXiv:1810.12488 [cs]},
}

@article{krizhevsky_learning_nodate,
    title = {Learning {Multiple} {Layers} of {Features} from {Tiny} {Images}},
    language = {en},
    author = {Krizhevsky, Alex},
}

@misc{clevert_fast_2016,
    title = {Fast and {Accurate} {Deep} {Network} {Learning} by {Exponential} {Linear} {Units} ({ELUs})},
    url = {http://arxiv.org/abs/1511.07289},
    doi = {10.48550/arXiv.1511.07289},
    urldate = {2026-05-06},
    publisher = {arXiv},
    author = {Clevert, Djork-Arné and Unterthiner, Thomas and Hochreiter, Sepp},
    month = feb,
    year = {2016},
    note = {arXiv:1511.07289 [cs]},
}

@article{jiang_recurrent_2026,
    title = {Recurrent {Network} {Expansion} for {Class} {Incremental} {Learning}},
    volume = {37},
    issn = {2162-2388},
    url = {https://ieeexplore.ieee.org/document/11142762/},
    doi = {10.1109/TNNLS.2025.3601373},
    number = {1},
    urldate = {2026-05-14},
    journal = {IEEE Transactions on Neural Networks and Learning Systems},
    author = {Jiang, Kai and Bai, Xueru and Zhou, Feng},
    month = jan,
    year = {2026},
    pages = {122--135},
}

@article{dong_dynamic_2026,
    title = {Dynamic expansion orthogonal network for class-incremental learning},
    volume = {341},
    issn = {09507051},
    url = {https://linkinghub.elsevier.com/retrieve/pii/S0950705126005940},
    doi = {10.1016/j.knosys.2026.115868},
    language = {en},
    urldate = {2026-05-14},
    journal = {Knowledge-Based Systems},
    author = {Dong, Mingda and Zhang, Zhizhong and Tan, Xin and Qiu, Jiling and Xie, Yuan},
    month = may,
    year = {2026},
    pages = {115868},
}

@article{li_insertion_2026,
    title = {{INSERTION}: {From} traditional incremental learning to open-world stream learning},
    volume = {176},
    issn = {00313203},
    shorttitle = {{INSERTION}},
    url = {https://linkinghub.elsevier.com/retrieve/pii/S0031320326001287},
    doi = {10.1016/j.patcog.2026.113163},
    language = {en},
    urldate = {2026-05-07},
    journal = {Pattern Recognition},
    author = {Li, Yanchao and Dou, Hongwei and Li, Guanxiao and Gao, Guangwei and Zhou, Huiyu},
    month = aug,
    year = {2026},
    pages = {113163},
}

@article{hussain_class-incremental_2026,
    title = {Class-incremental learning network for real-time anomaly recognition in surveillance environments},
    volume = {170},
    issn = {00313203},
    url = {https://linkinghub.elsevier.com/retrieve/pii/S0031320325007241},
    doi = {10.1016/j.patcog.2025.112064},
    language = {en},
    urldate = {2026-05-07},
    journal = {Pattern Recognition},
    author = {Hussain, Adnan and Ullah, Waseem and Khan, Noman and Khan, Zulfiqar Ahmad and Yar, Hikmat and Baik, Sung Wook},
    month = feb,
    year = {2026},
    pages = {112064},
}



\end{document}